# A novel testbed for investigating the impact of teleoperator dynamics on perceived environment dynamics

Mohit Singhala, *Student Member, IEEE,* and Jeremy D. Brown, *Member, IEEE*

*Abstract*— Human-in-the-loop telerobotic systems (HiLTS) are robotic tools designed to extend and in some circumstances improve the dexterous capabilities of the human operator in virtual and remote environments. Dexterous manipulation, however, depends on how well the telerobot is incorporated into the operator's sensorimotor control scheme. Empirical evidence suggests that haptic feedback can lead to improved dexterity. Unfortunately, haptic feedback can also introduce dynamics between the leader and follower of the telerobot that affect both stability and device performance. While concerted research effort has focused on masking these device dynamics or bypassing them altogether, it is not well understood how human operators incorporate these dynamics into their control scheme. We believe that to advance dexterous telerobotic manipulation, it is crucial to understand the process by which humans operators incorporate teleoperator dynamics and distinguish them from the dynamics of the environment. Key to this knowledge is an understanding of how advanced telerobotic architectures compare to the gold standard, the rigid mechanical teleoperators first introduced in the 1950's. In this manuscript, we present a teleoperator testbed that has reconfigurable transmissions between the leader and follower to change its dynamic behavior. The intent of this testbed is to investigate the effect of the teleoperator's dynamics on perception of and task performance in the remote/virtual environment. We describe the hardware and software components of the testbed and then demonstrate how the different teleoperator transmissions can lead to differences, sometimes significant, in the dynamics that would be felt by the operator when exploring the same environment.

## I. INTRODUCTION

Humans are capable of performing a myriad of highly dexterous tasks that require accurate estimation of environmental properties like stiffness, damping, and mass [1], [2]. This capability is developed over the course of one's life and refined with repeated manipulation of different objects in different environments under varying constraints. There are many tasks, however, where direct manipulation is either impossible, unsafe or inadequate. For many of these situations, the manipulation can be carried out through teleoperation. Examples of teleoperators include minimally invasive surgical robots, bomb disposal robots, and artificial robotic limbs. As with the natural limbs, however, dexterous manipulation through a teleoperator is only possible if the device's physical interactions with the environment are displayed to the operator in a manner consonant with their expectations.

In order to adequately perceive environmental dynamics, haptic sensory information originating from the energetic

Mohit Singhala and Jeremy D Brown are with the Department of Mechanical Engineering, Johns Hopkins University, Baltimore, MD, USA e-mail: mohit.singhala@jhu.edu

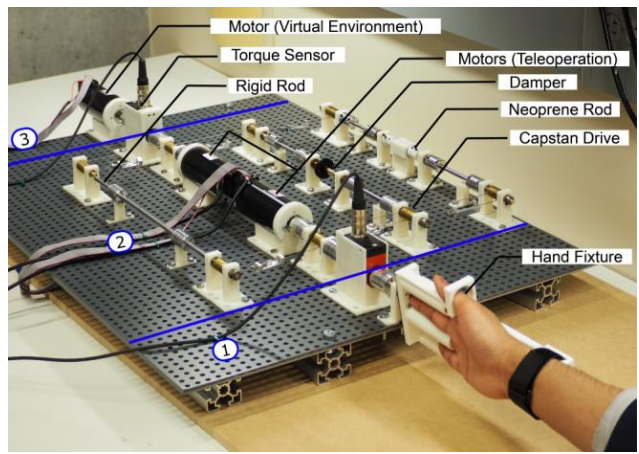

Fig. 1. Teleoperation testbed featuring three modules 1) the operator interface including the hand fixture in the neutral position, 2) a reconfigurable teleoperator with rigid, elastic, damped, and electromechanical transmissions, and 3) the environment which can be real or virtually rendered using a Maxon RE50 motor. The modules are coupled using Futek TRS600 torque sensors.

exchanges in direct manipulation is essential [2], [3]. However, introduction of a teleoperator between the operator and the environment significantly changes these energetic interactions, thereby altering the sensory information received by the operator. This phenomena has recently been explored in the context of effective impedance [4], [5], and illustrates how haptic device dynamics can amplify or attenuate the environment dynamics. In a bilateral force-reflecting teleoperator, device dynamics encapsulate both the internal dynamics of the leader and follower devices, and the transmission (mechanical or electromechanical) that connects them.

For electromechanical teleoperators, transparency has been a topic of considerable research over the last four decades [6], [7]. Transparency, by definition, requires that the operator's perceived impedance match that of the environment being explored through the teleoperator [8]. Thus, the transmission dynamics introduced by the teleoperator's closed-loop control architecture cannot distort this mechanical energy flow. Given the uncertainty associated with the dynamic state of the operator's limb(s), as well as the dynamic state of the environment, the pursuit of perfect transparency generally results in a tradeoff between system performance and control stability [3], [9]. Approaches such as passivity have attempted to improve teleoperator stability by monitoring the energy flow in the system [10]. Stable passive controllers, however, often lead to increased damping in the closed-loop dynamics, which decreases the telerobot's transparency and masks the dynamics of the environment felt by the operator.

Thus, true transparency represents a lofty ideal that has yet to be robustly achieved on physical systems.

As an alternative to direct force-reflection, there is a growing body of work evaluating the efficacy of cutaneous sensory substitution feedback as a means of replacing traditional kinesthetic cues [11]–[14]. The benefit of this approach is that it provides haptic information without directly affecting the operator's control over the teleoperator. To provide utility to the operators, these cutaneous cues need to be easily discernible and discriminable [15]. Likewise, they need to intuitively convey information about environment dynamics, a significant challenge given their non-colocated nature [2].

While passivity and sensory substitution hold promise, they are focused, respectively on compensating for the teleoperator transmission dynamics or bypassing them altogether. Neither approach, however, purposefully leverages the presence of the human in the control loop. From the perspective of tool usage, we know that humans can incorporate the dynamics of simple tools into their motor control scheme [16]–[18]. There is even evidence to suggest that this accommodation of device dynamics extends to teleoperators, with experienced operators relying more on internal models of the teleoperator than on external feedback [19]–[22]. It is also known that with training, operators are able to gain agency over the teleoperator [15].

While human capabilities of tool incorporation and agency are well known, they are not well understood. In particular, there are open questions regarding the impact of teleoperator dynamics on perception of the remote environment, and consequently, task performance in that environment. Likewise, it is unclear whether certain teleoperator dynamics are more or less disruptive than others. Finally, it is unknown to what extent environments with complex dynamics impact our understanding of perception through a teleoperator. From a historical perspective, the original mechanical teleoperators developed by Goertz [23] had inherent dynamics (mostly inertia). The same holds true for present-day mechanical teleoperators like body-powered prostheses and laparoscopic instruments. Thus, teleoperator users have always had to accommodate device dynamics of some form. As we continue to advance the field of telerobotics, it is imperative that we understand the human-centered consequences of altering the dynamic exchange between operator and environment.

To that end, we present a teleoperator testbed (Fig. 1) that is capable of simulating exploration with real and virtual environments through various mechanical and electromechanical teleoperator configurations. This reconfigurable teleoperator is designed to investigate how the leader-follower dynamics of a teleoperator affect perception of the remote environment and to what extent incorporation of these dynamics affects teleoperator agency and task performance. This system enables psychophysical and task-performance assessments of various phenomena related to the use of haptic sensory information in differing teleoperator use cases. In what follows, we provide a detailed description of the design of this testbed and an evaluation of its performance. We end with a discussion of some of its capabilities and detail potential future improvements.

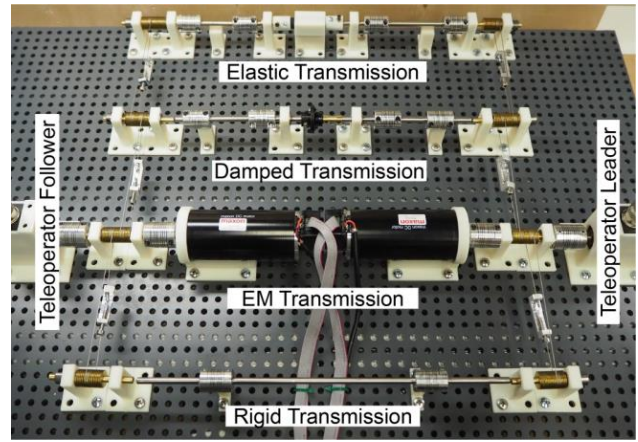

Fig. 2. Reconfigurable teleoperator with four transmissions 1) Rigid, 2) Electromechanical (EM), 3) Damped and 4) Elastic, connected in parallel using capstan drives to a common leader and a common follower.

## II. Design, Fabrication and Assembly

The following sections provide an overview of the testbed design, sensing and control. The testbed is a modular 1DoF closed-loop rotational device that can be reconfigured to function as a rigid mechanical teleoperator, a dynamic mechanical teleoperator with modular stiffness and damping, and an electromechanical teleoperator (see Fig. 1). This is achieved by selectively engaging a combination of four different transmission modules with a common user input and a common output to the environment.

### A. Module 1: Operator Interface

The operator interface serves as the connection between the operator and the teleoperator. The hand fixture (see Fig. 1) is designed for an alternating finger grip, enabling exploration via pronation and supination of the forearm. The interface is also compatible with similar hand-fixtures designed for radial/ulnar deviation and flexion/extension of the wrist. The system can also be configured to use conventional handle grips for exploration along any of the three degrees of freedom of the wrist.

### B. Module 2: Reconfigurable Teleoperator

The teleoperator comprises four distinct parallel transmissions (see Figure 2). All transmissions are connected via capstan drives and their input shafts rotate in the same direction as the hand fixture. This allows the four different transmissions to have one common input (leader) and one common output (follower). The different transmissions can be engaged and disengaged independently or simultaneously, to create various transmission dynamics and are described below.

1) Rigid Mechanical Transmission: A 316 stainless steel shaft is used to form a rigid transmission between the leader and the follower of the teleoperator. The shaft directly couples the input and the output with no torque or position scaling.
2) Elastic Transmission: A Neoprene rod (Shore 40A; 6.3 mm diameter; 50 mm length) is used as the elastic

element to couple the input and output capstan pulleys of this transmission. The rubber shaft is housed in a 3D printed fixture that surrounds the shaft in order to prevent buckling. Unlike standard torsional springs, the Neoprene rod allows bidirectional rotation (see Fig. 3a).

3) *Damping Transmission:* A gearless dual direction oil-filled rotary damper (9.45 mNm/rad/s) is used to couple the input and output capstan pulleys of this transmission. The damper has a minimum torque rating of 15 mNm (see Fig. 3b).

4) *Electromechanical Transmission:* This transmission uses two direct-drive Maxon RE50 (200 W) motors, each with a 3-channel 500 CPT HEDL encoder, to electromechanically couple the leader and the follower. The motors can be programmed to enable unilateral and bilateral teleoperators.

*Capstan Drives:* All capstan pulleys are machined from a stock of threaded Brass rod (5/8"-18), and have the same diameter, providing no torque or motion scaling. An 18-8 Stainless Steel, corrosion resistant, extra flexible wire rope (0.6 mm diameter) is used as the capstan wire. Unlike conventional capstan drives, where the capstan is often fixed to the larger arc, a custom machined U-column is used to complete the loop in this testbed. The columns are machined from lightweight 6061 Aluminium stock. Holes are drilled on either arms of the U-channel and the right arm is tapped to support a vented M2x0.4 mm screw. The capstan is first passed through the vented screw and clamped at the free end using a compression sleeve. The other end of capstan is then wound around the two capstan pulleys, passed through the hole in left of the U-channel, and clamped using another compression sleeve. Once both ends of the capstan are clamped to the U-channel, the screw is used to adjust the tension in the drive (see Fig. 3d).

All the capstan drives are set up in a parallel configuration on the input and output ends of the teleoperator. All input capstan pulleys rotate in the same direction, with the same angular velocity, regardless of the whether the transmission is actively engaged. Similarly, all output capstan pulleys rotate simultaneously, in the same direction with the same angular speed. The input and output shafts for each transmission are coupled to their respective capstan pulley using flexible servo couplings. Each transmissions can be independently engaged or disengaged by manually removing the servo coupling connecting the respective shafts and capstan pulleys (see Fig. 3e).

*Electromagnetic Coupling:* The manual coupling of transmissions through servo couplings is sufficient when performing tasks that use only one transmission. However, it may not be sufficient when conducting tasks that require rapid switching between transmissions on every trial, as required in most 2AFC psychophysical methods. To provide this capability, we designed an electromagnetic clutch module that can be easily coupled to the input and output drive shaft(s) of any mechanical transmission on the testbed. We use Ogura MIC 2.5NE electromagnetic clutches at the input and the output end of the transmission, to electronically engage and disengage it from the capstan drives. Custom 3D printed fixtures are used to couple the clutches to the drive shafts. These clutches can support torques of up to 250 mNm. These clutches were chosen based on their small form factor (see Fig. 3c).

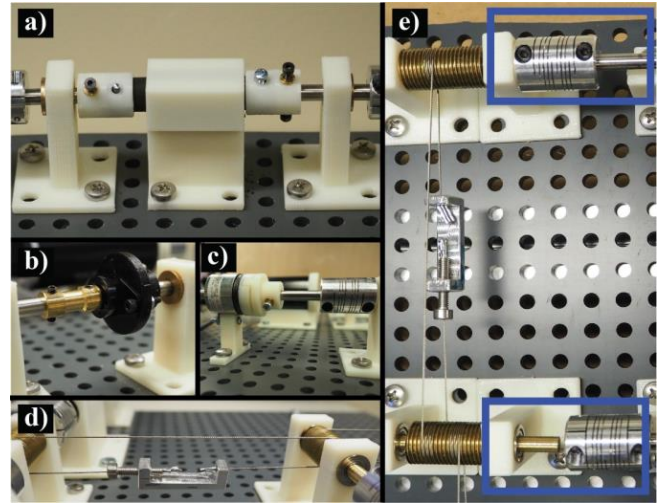

Fig. 3. a) Neoprene 40A shaft and housing, b) Rotary damper, c) Ogura electromagnetic clutch (only used when electronic switching is used instead of mechanical switching), d) close-up of the capstan drive and e) flexible servo couplings to manually engage or disengage the capstan pulleys and their respective drive shafts; the top highlighted area shows a shaft in the engaged position and the bottom highlighted area shows a shaft in the disengaged position.

### C. Module 3: Environment

The environment can be configured as any combination of real and virtual 1-DoF environments.

1) *Virtual Environments:* One direct-drive Maxon RE50 (200W) motor with a 3-channel 500 CPT HEDL encoder is used to render virtual environments with a peak torque of 467 mNm and a maximum continuous torque of 233 mNm.

2) *Real Environments:* The system enables rendering of a variety of real environments that may be either too complex to model or render in a stable manner using the Maxon motor. In our current iteration we can couple fluid containers of different shapes, volumes and weights to the output shaft.

### D. Couplings and Supports

The foundation of the testbed is a perforated high-strength PVC plastic sheet (6.35 mm thickness), supported by T-slotted aluminium extrusions at its base. Stainless steel rods (6 mm diameter; 316 SS) are used as the input and output shafts for all transmissions. Lightweight aluminium flexible servo couplings are used to couple the capstan pulleys to the drive shafts in order to accommodate any minor shaft misalignment. All shafts are supported using self-lubricated sleeve bearings and the capstan pulleys are supported by stainless steel ball bearings. Custom 3D-printed fixtures are used to couple the damping and elastic elements, and to create housings for the torque sensors and the motors. When a transmission is disengaged, its shafts, along with the servo couplings, rest on 3D printed supports that are fixed to the main support base for the device. The entire structure can therefore be reconfigured to accommodate differing spatial layouts and components.

*E. Data Acquisition and Control*

All four motors are controlled through a Quanser AMPAQ L4 linear current amplifier. Two Futek non-contact rotary torque sensors (TRS600) with a 5 Nm maximum torque capacity are coupled to the input and output shafts of the teleoperator. These sensors measure the torque applied by the operator to the teleoperator leader and the torque rendered by the environment to the teleoperator follower, respectively. All data acquisition and control operations are performed with a Quanser QPIDe DAQ, running at 1 Khz through a MATLAB/Simulink and QUARC interface. A monitor is used for visual stimuli, and a pair of Bose noise cancelling-headphones are available for user-studies. Custom Simulink blocks have been created to run standard psychophysical paradigms including Methods of Constant Stimuli and adaptive staircases.

## III. SYSTEM DEMONSTRATION

The testbed can be used to run a variety of perceptual and task performance experiments. In this section, we highlight the capabilities of the teleoperator testbed for performing studies assessing the impact of transmission dynamics on environment perception. Using the motor in the environment model, we render a virtual torsion spring of stiffness coefficient 1 mNm/deg. We then perform system identification to model the system at the teleoperator leader-operator interface, to understand how each of the five transmissions described below affect the dynamics that would be felt by the operator. We also consider the effect of these transmissions when no environment is rendered (i.e free space).

1) Rigid Transmission: Section II-B
2) Damped Transmission: Section II-B
3) Elastic Transmission: Section II-B
4) Combined Transmission: Damper and Elastic Transmissions (Section II-B) connected in parallel to form one dynamic transmission
5) Electromechanical Transmission: The leader and follower of the teleoperator are electromechanically coupled using two DC motors (Section II-B). The following position-position PD controller is used to render bilateral force-reflection between the leader and follower:

$$T_l(t) = K_p(\theta_f - \theta_l) + K_d(\dot{\theta}_f - \dot{\theta}_l)$$
$$T_f(t) = K_p(\theta_l - \theta_f) + K_d(\dot{\theta}_l - \dot{\theta}_f)$$

where $T_l$ is the torque applied to the leader motor, $T_f$ is the torque applied to the follower motor, $\theta_l/\dot{\theta}_l$ is the angular position/velocity of the leader, $\theta_f/\dot{\theta}_f$ is the angular position/velocity of the follower, $K_p$ is the proportional gain (0.05), and $K_d$ is the derivative gain (0.05).

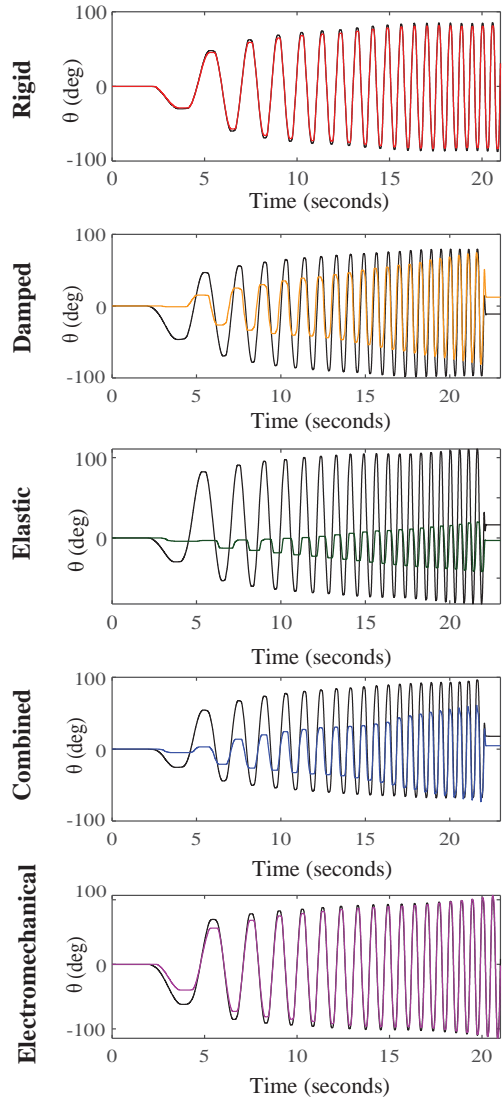

Fig. 4. Angular displacement in degrees, of the leader (black) and the follower (colored) for the five teleoperator configurations when exploring a 1m̈Nm/deg virtual spring.

*A. System Identification*

To obtain the torque/displacement relationship at the leader-operator interface, a position-based chirp signal was input to the teleoperator leader. The chirp oscillated between ± 90 degrees with respect to the neutral hand position (see Fig. 1) and linearly increased from 0.1–2 Hz in 20 seconds. The signal was designed to mimic the range of motion and nominal frequency of wrist motion expected from an operator [24]. Two chirp signals of the same range and amplitude were input to each system to obtain independent estimation and validation datasets. Angular displacements for the teleoperator leader (operator displacement) and teleoperator follower (environment displacement) were measured. Likewise, the torques applied at the teleoperator leader (operator torque) and the torque applied at the teleoperator follower (environment torque) were measured.

The Matlab 2018b System Identification Toolbox was used to obtain second-order models for the leader-operator system (experienced by the operator). Separate models were developed for the case where the teleoperator was connected to the 1 mNm/deg virtual spring and for the free space case (no environment present). Note, for the free-space case, the environment motor remained mechanically connected to the teleoperator, but was unpowered. This was done to ensure a proper measurement of output torque at the follower-environment interface, and to model the passive dynamics of the virtual environment motor.

TABLE I
TRANSFER FUNCTIONS AT THE LEADER-OPERATOR INTERFACE FOR
FREESPACE AND A VIRTUAL SPRING EXPLORED USING DIFFERENT
TELEOPERATOR TRANSMISSION CONFIGURATIONS

| Transmission | Freespace | Virtual Spring |
|---|---|---|
| Rigid | $\frac{3.41s+190.12}{s^2+6.51s+43.99}$ | $\frac{-2.04s+163.26}{s^2+4.19s+195.11}$ |
| Damped | $\frac{-2.25s+119.61}{s^2+6.65s+76.12}$ | $\frac{-0.18s+72.52}{s^2+4.71s+97.64}$ |
| Elastic | $\frac{0.02s+127.95}{s^2+9.83s+98.73}$ | $\frac{-0.75s+179.63}{s^2+7.09s+133.42}$ |
| Combined | $\frac{-0.83s+64.91}{s^2+6.26s+90.33}$ | $\frac{-1.7s+95.32}{s^2+7.55s+130}$ |
| Electromechanical | $\frac{-2.7s+174.31}{s^2+6.08s+48.88}$ | $\frac{-2.82s+172.09}{s^2+6.33s+48.55}$ |

### B. Results

We observed no appreciable slip in the capstan drives, as demonstrated in the tracking performance of the follower when using the rigid transmission (Fig. 4 previous page), for the entire range of motion, across the frequency spectrum. We also share the step response and bode diagrams for all identified systems (see Fig. 5 next page). All systems obtained a greater than 99% fit when fitting a second order system, treating torque (in mNm) as the input and displacement (in degrees) as the output. All identified systems passed the test for independence with the default parameters of 20 samples amd a 95% confidence interval. The transfer functions for all the systems are shared in Table I:

## IV. DISCUSSION

### A. System Identification

The results demonstrate how the dynamics felt at the leader-operator interface can be significantly different from the rendered dynamics. Depending on dynamics of the teleoperator transmission, the rendered dynamics may get significantly attenuated or completely masked by the teleoperator dynamics. While the dynamic response highlighted in Section III-B is unique to the specific hardware and software parameters of our current system, the following observations regarding system models for the five different teleoperation configurations can be generalized:

1) Rigid transmission: We found that the follower displacement tracks the leader displacement well, with negligible slip, confirming the efficiency of the capstan drives (see Fig. 4). We observed that explorations made through the rigid transmission produced a system that most closely resembles the environment (free space and virtual spring). This is to be expected as the inherent dynamics of a rigid teleoperator can predominantly be modeled as inertia with less damping compared to the other transmissions.

2) Damped transmission: We observed that at low frequencies the follower displacement does not track the leader displacement well (see Fig. 4). However, at high frequencies, leader-follower tracking improves, exhibiting behavior similar to that of the rigid transmission. This behavior is expected from a damped transmission. We also observed that explorations made through the damped transmission produced a system that resembles the environment (free space and virtual spring), however the overall amplitude of the system response is dampened compared to the rigid rod.

3) Elastic transmission: We found that the follower displacement does not track the leader displacement well (see Fig. 4). We also observed that explorations made through the elastic transmission produced a system that resembles the system with the damped transmission. Both of these behaviors can be explained by the high compliance (low stiffness) of the Neoprene 40A rod used as the elastic element, which likely possess some appreciable viscoelastic properties.

4) Combined transmission: We observed that the combined system exhibited leader-follower tracking behavior that most closely resembles the damped transmission. This can be attributed to the low compliance of the elastic transmission, which contributes little to the effective dynamics of the combined transmission. We also observed that explorations made through the combined transmission produced a system that attenuated the environment dynamics the most compared to the other mechanical transmissions. This is likely due to the added damping, elasticity, and inertia in the combined system.

5) Electromechanical transmission: We observed that the electromechanical transmission exhibited leader-follower tracking behavior that was the closest to the rigid transmission (see Fig. 4). At the same time, however, we observed that explorations made through the electromechanical transmission produced a system that was almost identical in response between the free space and virtual spring environments. It is likely the case that the damping coefficient used for stable control were dominant in defining system behavior and the virtual spring rendered at the environment had a negligible effect on the effective dynamics at the leader-operator interface.

### B. Design Considerations

The system can support over 180 degrees of angular displacement input, covering the complete range of motion for a operator's wrist . All the capstan drives are connected in parallel with a common leader and a common follower to maintain the modularity of the system. By setting up the capstans in parallel, they do not have to be unwound and rewound each time the teleoperator configuration is changed. We believe this design decision is justified, considering the tradeoff between the added inertia from three capstan drives versus the loss of the modular function of the testbed.

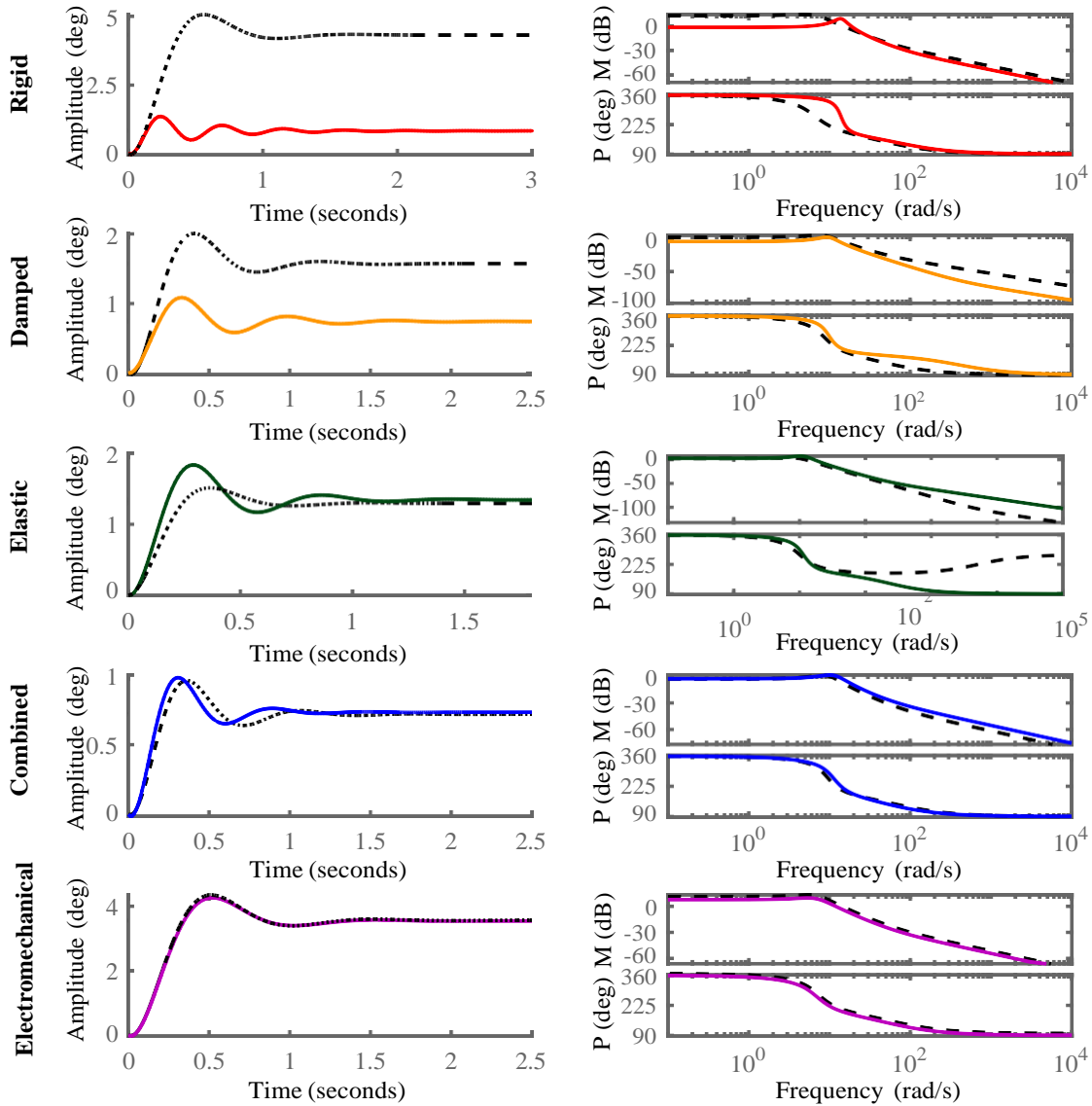

Fig. 5. Step response and bode plots for systmes rendered at the Teleoperator leader-operator interface, with Torque as input (mNm) and Angular Dipslacement as output (deg), when exploring a 1 mNm/deg virtual spring and freespace (FS) for the five different teleoperator configurations. The freespace response is represented using the black dashed line. "P" denotes Phase and "M" denotes Magnitude for the bode plots

We mount the entire testbed on a perforated PVC sheet (6.35 mm) thick, supported by T-slotted Aluminium extrusions. All 3D printed parts use a common base design to enable easy fixation and removal from the PVC base without need of additional machining. All active elements of the different teleoperation transmissions are mounted on an independent set of bearings such that the properties of each transmission can be changed without disassembling the entire transmission. Low-weight flexible couplings are used for all shaft couplings to account for any misalignment and minimize inertia from the transmission.

A Neoprene rod is used as the elastic element as conventional torsion springs are designed to be displaced in only the direction in which they are wound. Neoprene (Shore 40A) was chosen as a 50 mm section of 6.35 mm diameter Neoprene does not buckle under it's own weight. The shore hardness of Neoprene rubber also exhibits a linear correlation with its Young's modulus (1.69 MPa for Shore 40A hardness), thus acting as an ideal bidirectional stiffness element.

We also note from the free space system models (see Fig. 5) that all teleoperator transmissions show inherent compliance. We believe that the source of this compliance is most likely associated with the capstan drives, motors, and supports used in the system. We will investigate this assumption in future design improvements. At present, however, this compliance does not affect the functionality of the testbed.

### C. Future Improvements

The testbed and the modules in its current iteration are focused on studying the effects of teleoperator dynamics on perception of remote environment and evaluation of task performance in that environment. The testbed can also be configured to study other aspects of haptic perception of teleoperated environments, including modeling of the complex environment-follower system using the torque sensor and encoders coupled to the follower. The testbed serves as a validation setup for investigating the performance of novel controllers and comparing that performance to the rigid teleoperatoer configuration. We are currently investigating the use of this testbed to model real non-linear systems, and replace them with virtual environments. The closed-loop nature of the testbed creates the potential to create data-driven models of systems that would otherwise be difficult to model. We believe that this functionality would be particularly useful for psychophysical investigations which require a wide range of stimuli. The environment module can also be expanded to test the effects of preturbations on perception of real environments by connecting real and virtual environments in series. Additional lines of inquiry include investigating the effects of delay in real-time exploration of remote environments. Furthermore, this system could one day prove useful as an educational tool on dynamics and teleoperation. In future design iterations, we will also aim to minimize the inherent compliance present in each of the teleoperation transmissions.

## V. Conclusion

In this manuscript, we presented a modular teleoperation testbed that can be configured with rigid, dynamic, or electromechanical transmissions. We highlighted the engineering design of the testbed and demonstrated its capabilities by means of system identification. We showcased how the teleoperator transmission dynamics can mask the environment dynamics resulting in the system sensed by the operator feeling different from the system rendered as the environment. Given the human ability to incorporate teleoperators with training, it is expected that the teleoperator dynamics themselves may be modeled separately, and hence, affect the perception of the environment dynamics. To that effect, we believe that this testbed will serve as a useful tool to study how the teleoperator dynamics affect the perceived dynamics of a remote environment, and consequently, how it affects task performance in that environment. The testbed also provides an opportunity for comparing advanced teleoperator control approaches to the gold standard, the rigid mechanical teleoperator.

## Acknowledgment

This material is based upon work supported by the National Science Foundation under NSF Grant #1910939. We thank Dr. Evan Pezent for his assistance with the design of the capstan drives.